\documentclass[letterpaper, 10 pt, conference]{ieeeconf}  

\IEEEoverridecommandlockouts                              
                                                                                                       
\usepackage{booktabs}

\usepackage{cite}
\usepackage{amsmath,amssymb,amsfonts}
\usepackage{algorithmic}
\usepackage{graphicx}
\usepackage{textcomp}
\usepackage{xcolor}
\usepackage{subfig}

\usepackage{textcomp}

\usepackage{fancyhdr}
\usepackage[ruled,vlined]{algorithm2e}

\usepackage{amsmath}
\usepackage{fancyhdr,graphicx,amsmath,amssymb}
\usepackage[ruled,vlined]{algorithm2e}
\usepackage{bm}
\usepackage{xcolor}

\usepackage{multirow} 

\include{pythonlisting}

\def\BibTeX{{\rm B\kern-.05em{\sc i\kern-.025em b}\kern-.08em
    T\kern-.1667em\lower.7ex\hbox{E}\kern-.125emX}}

\usepackage{bm} 
\usepackage{color} 

\usepackage{tikz}
\usetikzlibrary{decorations.pathmorphing} 
\usetikzlibrary{matrix} 
\usetikzlibrary{arrows} 
\usetikzlibrary{calc} 
\usepackage{tikzscale}

\tikzstyle{block} = [draw,rectangle,thick,minimum height=2em,minimum width=2em]
\tikzstyle{sum} = [draw,circle,inner sep=0mm,minimum size=2mm]
\tikzstyle{connector} = [->,thick]
\tikzstyle{line} = [thick]
\tikzstyle{branch} = [circle,inner sep=0pt,minimum size=1mm,fill=black,draw=black]
\tikzstyle{guide} = []
\tikzstyle{snakeline} = [connector, decorate, decoration={pre length=0.2cm,
                         post length=0.2cm, snake, amplitude=.4mm,
                         segment length=2mm},thick, magenta, ->]

\usepackage{etoolbox}
\makeatletter
\patchcmd{\@makecaption}
  {\scshape}
  {}
  {}
  {}
\makeatletter
\patchcmd{\@makecaption}
  {\\}
  {.\ }
  {}
  {}
\makeatother
\def\tablename{Table}
\def\tablename{Table}

\begin{document}

\title{Survivable Hyper-Redundant Robotic Arm with Bayesian Policy Morphing\\
}

\author{
Sayyed Jaffar Ali Raza,
Apan Dastider 
and
Mingjie Lin
}

%


\maketitle

\begin{abstract}
%
In this paper we present a Bayesian reinforcement learning framework that allows robotic manipulators to adaptively recover from random mechanical failures autonomously, hence being survivable. To this end, we formulate the framework of {\em Bayesian Policy Morphing} (BPM) 
that enables a robot agent to self-modify its learned policy after the diminution of its maneuvering dimensionality. We build upon existing actor-critic framework, and extend it to perform policy gradient updates as posterior learning, taking past policy updates as prior distributions. We show that policy search, in the direction biased by prior experience, significantly improves learning efficiency in terms of sampling requirements. We demonstrate our results on an 8-DOF robotic arm with our algorithm of BPM, while intentionally disabling random joints with different damage types like unresponsive joints, constant offset errors and angular imprecision. Our results have shown that, even with physical damages, the robotic arm can still successfully maintain its functionality to accurately locate and grasp a given target object.

\end{abstract}


\section{Introduction}
\label{sec:intro}


%

%
%



%
%

Cooperative working at irregular environment
stipulates human beings to be intelligent, adaptive, and safe within their environments. Humans and animals can also adapt to physical challenges (ex. amputation, compromised locomotion etc.) which is analogous to fault tolerance mechanism in robotics ~\cite{Goodrich2007,Schmitt2019a}. Performance of a robot agent can be significantly compromised if it experience even slight physical damage while operating in an irregular environment.
%
On the other hand, animals not only can react differently to different surroundings,
but also sustain their maneuvering capability even 
after serious bodily injuries, thus highly adaptive.
As of today, 
most robotic systems 
do not possess a high degree of functional tolerance,
mainly because 
it is quite challenging for robotic agents,
in response to sudden mechanical malfunctions and environment changes,
to autonomously self-modify their own control policies 
and 
quickly generate qualitatively different compensatory behaviors,
except resorting to limited predefined contingency plans.

\begin{figure}[htbp]
	\centering
	\includegraphics[width=\linewidth]{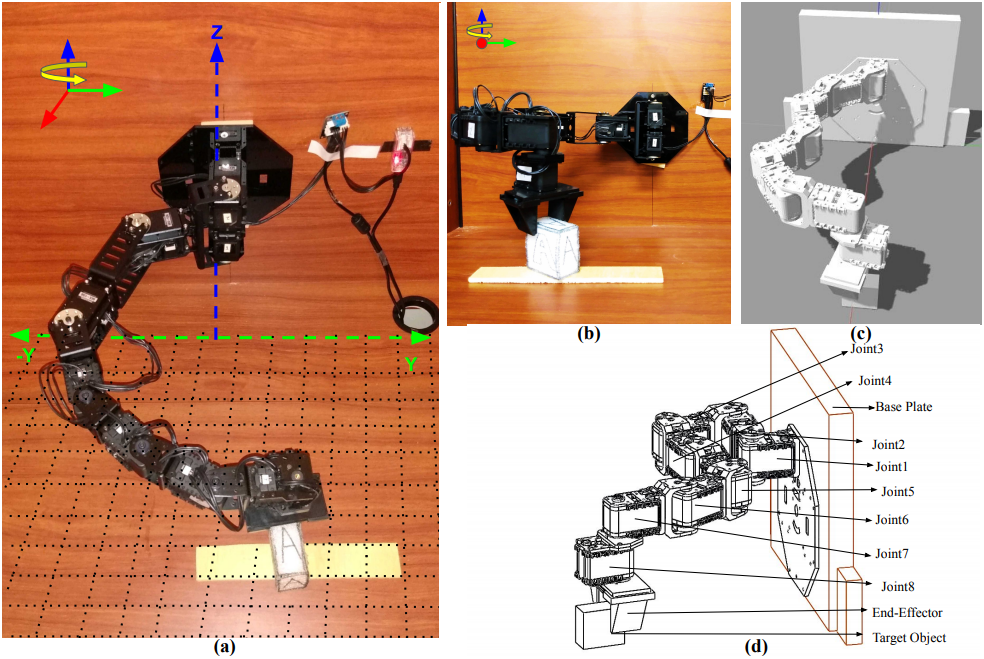}
	\caption{
(a) Physical design for redundant manipulator. 
(b) Front view showing reach of arm's end-effector. 
(c) Exoskeleton in Gazebo-ROS simulation used during training.
(d) 3D CAD model of overall hyper-redundant design.
}%
	\label{fig:approach}
\end{figure}

There are at least three fundamental challenges in making survivable robotics.
First, how to abstract out an existing motion policy into a useable form that 
can bootstrap policy exploration later on.
Second, to ensure the timeliness of response,
a survivable robotic agent has to compute an effective control policy that meets 
a real-time learning performance.
Third, although Bayesian algorithms, based on apriori data, can incrementally predict posterior
distributions with probabilistic inference,
posterior prediction becomes inconsistent, if the learning model is sensitive to small parametric perturbations. This is typically
the case for reinforcement learning (RL), specifically when learning a policy
in continuous spaces with stochastic dynamics. 
To alleviate all these challenges under a robotic setting, 
this paper formulates the problem of {\em Bayesian Policy Morphing} (BPM).
Abstractly, given a learned policy $\pi$ for a fixed robotic agent, 
BPM aims at quickly adapting to 
sudden changes in system dynamics or robotic agent itself. 
For example, a trained robot unexpectedly losses certain portion of its mobility or the 
unforeseen occurrence of obstacles or hazards.
Mathematically, all these scenarios can be formally abstracted as the losses in 
state space $\mathcal{S}$ or action space $\mathcal{A}$. 
As such, the goal of our BPM is to compute a new optimal policy $\pi'$ 
considering the changes in $\mathcal{S}$ and
$\mathcal{A}$, while exploiting the previously learned policy $\pi$.
%
With such a Bayesian-based learning strategy, our BPM methodology enables 
a hyper-redundant robotic arm to recover its functionality even after multiple joints are either stuck or damaged.

Our key idea is to integrate the Bayesian learning framework with
the deep reinforcement learning while systematically modifying the behavior of an agent as well as providing an efficient estimation framework for an agent to learn a policy by both taking actions---exploring state-values to enhance ground belief, and predicting instead of acting---inferring future value distributions
instead of sampling~\cite{bpg,bpr},
hence exploiting its base policy as much as possible.

\begin{figure}[htbp]
	\centering
	\includegraphics[width=0.8\linewidth]{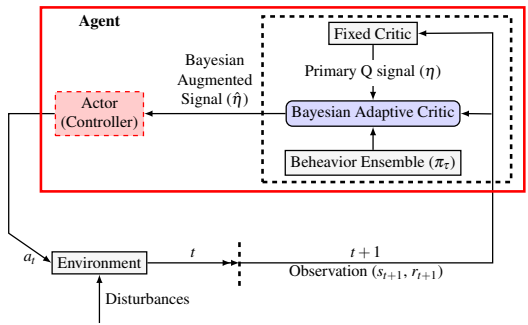}
	\caption{BPM idea and its difference from standard RL.}%
	\label{fig:overall}
\end{figure}%

Fundamentally, as shown in Fig.~\ref{fig:overall},
the theoretical basis of our BPM methodology is the standard model-free off-policy deep RL~\cite{actor_critic,Gu2017}
that uses deterministic policy gradient~\cite{Lillicrap2016} as policy updating method. 

At a very high level, our overall learning strategy is almost identical, 
to the standard Actor-Critic RL setting, except that the BPM agent (red box in Fig.~\ref{fig:overall}) consists of a {\em Bayesian Adaptive Critic} that 
combines the inputs from both a fixed critic and its behavior ensemble,
and subsequently generates the Bayesian augmented signal.

More specifically, as depicted in Fig.~\ref{fig:overall}, 
our BPM algorithm consists of two technical ingredients that differ from the standard RL framework:
(1) 
Generate posterior distribution for updating successor policy parameters by inducing bias by exploiting past policy parameters as latent behavioral representation; 
(2)
similarity heuristic to perform commonality estimate of apriori distribution
with respect to on-going policy distribution. 
We believe that our BPM problem formulation can be widely useful in many empirical settings. 
%

{\bf Contributions}. 
Our BPM algorithm 
exploits a given prior policy distribution to bootstrap parameter learning,
after the target robotic agent unexpectedly suffers a significant loss in 
its state and/or action space, such as malfunction or stuck-at errors 
in a robot's servo joints.  
Our paper claims the following contributions:

\begin{enumerate}

\item Instead of learning parameters from scratch, integrate previously learned policy distribution as behavior ensemble with newly received observations in order to incrementally synthesize a new policy parameters.

\item Our proposed BPM algorithm can generalize to tackle unexpected mechanical damages in robotic system through quickly evolving its prior motion policy at a high RL convergence rate.
We consider three types of mechanical malfunctions that include: 
1) complete joint frozen,
2) joint always offset by 30 degrees,
and 
3) joint has stochastic rotational inaccuracy.
In addition, each malfunction type has the number of affected joints ranging from 1 to 4 (see Fig.~\ref{fig:boxplots}).

\item We have demonstrated the effectiveness of our proposed BPM algorithm in both simulation and
experimental environments. Our experimental robotic agent has significantly outperformed,
in terms of learning rate and final positional precision.
a baseline with the state-of-the-art DDPG-trained robot~\cite{Lillicrap2016} 
augmented by continuous control scheme~\cite{Ota2019}.

\end{enumerate}


\section{Related Work}


\subsection{Deep Reinforcement Learning in Robotics}

Deep RL~\cite{Barto2019} 
theoreticizes a comprehensive and analytical framework, 
with minimal human intervention,
for robotics to achieve sophisticated and hard-to-engineer behaviors
even with 
insufficient domain knowledge of system dynamics~\cite{Gu2017,Lillicrap2016}.
As such, deep RL finds wide applications spanning from robotic manipulation
and autonomous driving.
However, in real-world robotic applications,
directly applying deep RL to robotics often suffers 
from curse of dimensionality, curse of real-world samples, and curse of under-modeling~\cite{Kober2013}.
Only quite recently, 
deep RL algorithms based on off-policy training of deep
Q-functions have been demonstrated to handle complex 3D manipulation tasks and learn
policies efficiently enough to train real physical robots~\cite{raza2019constructive,Lillicrap2016,skh}.
%

%
%
\subsection{Bayesian Reinforcement Learning}

Bayesian formalism, 
through incorporating prior information with inference, 
offers an
elegant approach 
to account for
prior knowledge and observational uncertainty while 
effectively trading between exploration and exploitation
in RL.
%
Unfortunately in modern robotics,
the prior information and system model 
can rarely be expressed as a parameterized Markov model~\cite{Ghavamzadeh2016},
therefore we will focus exclusively 
on Bayesian methods for model-free RL, where priors are expressed over the
value function or policy class.
Specifically,
the Gaussian process temporal difference (GPTD)~\cite{Enge2005},
a value-based Bayesian RL, 
employs the Bayesian methodology to infer a
posterior distribution over value functions conditioned on the state-reward
trajectory observed in learning 
with an accuracy measure, value estimation variance, as a byproduct.
Recently, 
the Bayesian policy gradient (BPG) algorithms~\cite{Ghavamzadeh2016}, 
a representative policy-based Bayesian RL technique,
proposed to maintain a class of smoothly parameterized stochastic policies
$\{ \pi ( \cdot | s; \boldsymbol{\theta}), s \in \mathcal{S}, \boldsymbol{\theta}\in{\Theta} \}$,
and update the policy parameters $\boldsymbol{\theta}$ in the direction
of the estimated gradient of a performance measure.
%
Finally, the Bayesian actor-critic (BAC) algorithms~\cite{Ghavamzadeh2007}
innovatively applied the Bayesian quadrature (BQ) machinery~\cite{Ghahramani2002} to
the policy gradient in order to reduce the
variance in estimating policy gradient.

\subsection{Bayesian Reinforcement Learning for Robotics}

For statically constrained 
working environment, 
utilizing deep RL-based algorithm to optimize motion trajectory 
for robotic systems with unknown dynamics 
has been investigated previously.
For example, 
a trajectory-centric RL (Guided Policy
Search) approach~\cite{Ota2019}
was proposed to intelligently reshape
reward functions based on trajectories computed by RRT~\cite{Casalino2019}.
%
Only quite recently,
researchers started leveraging deep RL to autonomously constructing
sequential manipulation policy  
with unknown robot dynamics and time-variant environments. 
For example, studies~\cite{Schmitt2019} and~\cite{Schmitt2019a} 
investigated, with a dual-robot setup similar to ours,
automatically deriving constraint-based
controllers and using them as steering functions in a kinodynamic manipulation
planner with collision avoidance.
Their central idea is to build multiple policy priors that tackle different environments,
and dynamically select one,
guided by Bayesian learning framework, 
to react on-line to disturbances.
Fundamentally, both studies shared with the meta-learning 
proposed in~\cite{Frans2018} the key idea that 
the optimal policy can be computed through intelligently sampling from 
a meta-policy or a policy pool trained for multiple objectives or environments.

{\bf Key BPM differences}. 
Whereas the state-of-the-art on Bayesian RL~\cite{Ghavamzadeh2016,Ghavamzadeh2007,Ghavamzadeh2015}
has utilized Bayesian learning to more accurately estimate value function or policy gradient,
our BPM algorithm exploits Bayesian learning to principlely guide 
policy search in the direction biased by prior experience. 
In terms of problem setup, 
while studies such as~\cite{Ota2019,Schmitt2019,Schmitt2019a,Frans2018} 
considered robotic control under static environment, 
our BPM work instead focuses on 
adaptively synthesizing optimized motion policy when subjected to sudden changes in state and action spaces, 
e.g., mechanical malfunctions or failures.

\section{Proposed Approach}\label{sec:approach}
\begin{figure}[htbp]
	\centering
	\includegraphics[width=\linewidth]{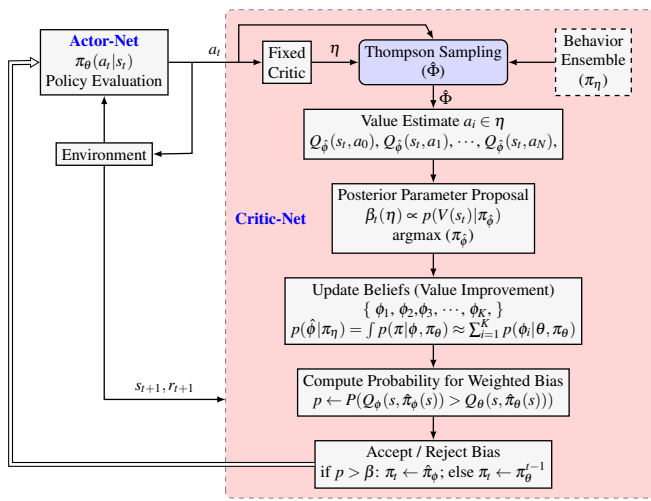}
	\caption{Overall BPM Algorithm.}%
	\label{fig:algo}
\end{figure}

Our BPM methodology has fundamentally been built upon 
the recent work of DDPG~\cite{Lillicrap2016, a-c-algo}
and the asynchronous off-policy updating strategy~\cite{Gu2017},
but with significant extensions of Bayesian learning (See Fig.~\ref{fig:overall})
and augmented by continuous control scheme~\cite{Ota2019}. 
Specifically,
our BPM combines probabilistic estimation
using behavior ensemble and Thompson Sampling \cite{thompsonSampling} to
sample unknown MDPs (see Fig.~\ref{fig:algo}).
In addition, our BPM
utilizes deep function approximators 
that can learn policies in high-dimensional and continuous action spaces, 
which is crucial for robotic control.
Furthermore, our BPM stabilizes its non-linear deep approximation through
relay buffer 
sampling 
and batch normalization,
both off-policy.

\subsection{Confidence Metric Representation}

We formalize the problem as agent based learning as shown in Fig.~\ref{fig:algo}. 
The agent constitutes its
interaction with environment as an MDP, such that at any given state $s$ at
timestep $t$, the agent follows a policy $\pi(s_{t})$ to choose action
$\pi(s_{t})\rightarrow a_{t}$.  Each state has a value, that is expectation of
discounted cumulative sum of rewards
\begin{equation}
 V_{\pi}(s_{t}) = \mathbb{E} \left[\arg\max_{a}\left(\sum_{k=0}^{N} \gamma^{k} r_{t+k+1}\,|\, s_{t},a_{t}=\pi(s_{t})\right)\right],
\end{equation}
where $0<\gamma<1$ is the discount factor. For higher dimensions, the policy
can be parameterized with vector-values $\theta$ to approximate model. The
performance of policy $\pi$ is defined as probability distribution over
expected utility value $P (R | \theta,\pi,\eta)$, where $\theta$
is policy parameter for model $\eta$ and $R$ is cumulative reward. We
explicitly represent model dynamics as $\eta$ to signify that $\theta$ is
conditioned over stationary model representations. When the model
representations are varied from $\eta\rightarrow\eta^{\prime}$, the parameter
$\theta$ would need to be readjusted by resampling state-action values to
comply with variations, 
which is computationally expensive task to perform
every time the model dynamics are changed.

\subsection{Probabilistic Sampling of Behavior Ensemble}

Given a behavior model $\eta$ as a probability distribution, 
we draw samples that strongly correlate with on-going distribution $\hat{\eta}$ of
the agent, 
while maximizing expected reward as well as acting
as randomly as possible to balance the exploitation of behavior policy in those
states whose utility is irrelevant (unknown) for current distribution
$\hat{\eta}$. We use Thompson sampling to draw probabilistic sample
distributions my modeling posterior reward distribution of each state-action
variable.

Thompson Sampling, instead of maintaining a fixed-point estimate, maintains a
distribution over ensemble weights and consequently over values of the state
space. The samples are contextualized as set of parametric weights for
$\hat{\eta}$ that can be forward-passed to the critic network to generate
resulting Q values for each state-action combination. This technique is well
known in Multi-Arm bandit problems where each bandit carries unique
distributions and agent only has a single independent prior belief. We extend
this setting over a continuous control agent setting where an agent is provided
with a behavior ensemble as its belief, and uses Bayes' rule to estimate posterior
distributions.

\subsection{Sample Reduction using Apriori Morphing}

In order to reduce sampling requirements, our BPM defines the confidence metric
$\beta(\eta)$ as probability distribution that measures the extent of match
between representation $\eta$ (already seen) and representation $\eta^{\prime}$
(currently faced).

\begin{equation}
    \beta_{t}(\eta) \propto \operatorname{P}\left(V (s_{t}) | \theta,\pi_{t},\eta\right)\beta_{t-1}(\eta)
    \label{eq:2:confidence}
\end{equation}

When learning unknown representation $\eta^{\prime}$, $\eta$ is taken as
behavior ensemble that initializes confidence metric with a prior probability
using Bayes' rule at timestep $t_{0}$. $\beta$ metric identifies the confidence
as belief over Bayesian apriori, which is then used as bias kernel to use
weighted hypothesized increment on current policy optimization. If more than
one model representations are accessible, than the confidence parameter can be
represented simply as maximum expectation: 

\begin{equation}
    \eta\leftarrow\max_{\eta}\sum_{i=0}\beta(\eta_{i})\operatorname{E}\left[R|\eta_{i},\pi\right], \quad \eta_{i}\in\boldsymbol{\bar{\eta}}
\end{equation}

where $\boldsymbol{\bar{\eta}}$ represents set of all representations. We use
only single model representation for our experimentation, to emphasize on
adaptability aspect of BPM (instead of multi-policy) as major contribution of
this work. 

\subsection{Bayesian Policy Morphing}

Once the confidence distribution is computed, the BPM proposes bias parameter
$\hat{\eta}\approx P(\hat{\eta}|\eta,\eta^{\prime})$ by taking joint
probabilities of two available representations. Any improvement to expected
utility for currently facing representation would use confidence as kernel. We
can write optimal policy improvement function as:

\begin{equation}
    \pi^{*}=\arg\max_{\eta} \int^{+\infty}\sum_{\eta}\beta(\eta)\operatorname{P}\left(V (S) | \theta,\pi,\eta\right) d \theta
\end{equation}

The optimal policy is conditioned over confidence metric within the critic
framework. And the accuracy of confidence is highly dependent on the
performance distribution $\operatorname{P(R|\theta,\pi,\eta)}$ which has both
policy parameter $\theta$ and confidence metric $\eta$  as conditional
parameters. The conditional probability makes confidence distribution agnostic
to myopic value estimates that can arise if $\gamma$ is set too close to $0$.

\begin{algorithm}
\scriptsize
   \caption{Picker Algorithm}
\label{algo:pick}
\SetAlgoLined
  \SetKwData{Left}{left}
  \SetKwData{Up}{up}
  \SetKwFunction{FindCompress}{FindCompress}
  \SetKwInOut{Input}{input}
  \SetKwInOut{Output}{output}
  \textbf{Require:}
  \\ \quad $\pi_\tau$ : Behavior Ensemble
  \\ \quad $S_t$ : Current Observation
  \\ \quad $T$ : $\#$ of allowed steps
  \BlankLine
 \For{$episode=\{1,2,...,k \}$}{
 \For{$t=\{1,2,...,T\}$}{
 $a_t \leftarrow$ obtain from $\pi_\theta(a_t|s_t) $
 \\ $\eta \leftarrow FixedCritic(a_t,S_T,\theta)$
 \\ $\hat{\phi} \leftarrow TS(\eta|\pi_\eta)$
 \\Update:
  \\\quad \text{\scriptsize{Value Estimate: $Q_{\hat{\phi}}(s_t,a_0),Q_{\hat{\phi}}(s_t,a_1),...Q_{\hat{\phi}}(s_t,a_T)$}}
 \\\quad \text{\scriptsize{Posterior Param $\leftarrow \beta_{t}(\eta) \propto P(V_{s_{t}}|\pi_{\hat{\phi}})$ \CommentSty{\scriptsize{\# eq. 2}}}}
 \\ \quad \text{\scriptsize{Value Improvement $\leftarrow \rho(\hat{\phi}|\pi_{\tau}) \approx$ $\sum_{i}P(\phi_{i}|\theta,\pi_\theta)$}}
 \\ \CommentSty{\scriptsize{\# Get probability of expected value}}
 \\$\rho \leftarrow P[Q_{\phi}(s,\hat{\pi_{\phi}}(s))>Q_{\theta}(s,\pi_{\theta}(s))]$
 \\ \uIf{$\rho>\beta$}{
   \CommentSty{CNTRL}: accept proposed bias
 \\ choose $a_t \leftarrow \mathcal{A}_\phi$
 }
\uElse{
$V(s_t)\leftarrow FixedCritic$
 }
 \CommentSty{CNTRL} update:
 \\ \quad Loss $\phi\leftarrow\phi - \Delta_{\phi}L_{critic}(\phi)$
  \\ \quad Loss $\theta\leftarrow\theta - \Delta_{\theta}L_{critic}(\theta)$
  \\ \quad Value Update$\leftarrow V(s_t)$
 }
 return $\hat{\eta}$
 }
 
\end{algorithm}

\subsection{Accept/Reject Controller}

Lastly, the overall setting of BPM also rectifies the bias induction caused by
inaccurate posterior estimations. Equation: \ref{eq:2:confidence} shows the
confidence as most appropriate similarity of $\eta^{\prime}$ with $\eta$. This
can be simply rewritten as optimal belief at any given time $t$

\begin{equation}
    \left. \beta^{*}(\tau)=\arg\max_{\eta}{\operatorname{P}(V_{t})}\beta_{t}(\eta)\right.
\end{equation}

The confidence $\beta^{*}(\tau)$ can be considered as unique solution if and
only if $\beta^{*}(\tau)>\beta^{*}(\tau)$. This condition can also be defined
in proportion to agent's current distribution as
\begin{equation}
    \beta^{*}_{t} \equiv \beta^{*}_{t}(\eta|r_{t},\pi_{t},\theta_{t})\propto\operatorname{P}(R|\pi_{t},\theta_{t}, \eta). 
\end{equation}
The belief proposals could be accepted or rejected based on this sanity check
to rectify the confidence to only maximize the utility. If the confidence
results in detrimental update, the BAP controller would reject the bias
proposal (see Fig.~\ref{fig:algo}). The overall flow of BPM
is outlined as pseudo-algorithm in Algorithm~\ref{algo:pick}.

\section{System Overview}
\label{sec:system}

This section describes our experimentation system
comprising of both hardware and simulation platforms in order to produce experiment results
as presented in Section~\ref{sec:res}.

\subsection{Experimental Setup}

As shown in Fig.~\ref{fig:approach},
we used a custom-built 8-DOF hyper-redundant robot arm 
mounted on the vertical frame,
which faces a horizontal table-top
plane constituting the workspace for its manipulation reach. 
The first 3 DOFs of our robotic arm are realized 
with three Robotis Dynamixel MX-106 servo motors
with high torque and robust structural durability, 
while its next 4 DOFs are controlled with AX-12A motors, 
and its last DOF is supplied by a gripper as the end-effector 
to pick an object placed in its workspace.
Furthermore, all these actuators are controlled through a
daisy-chained command control system. 

To conduct mechanical control,
we integrated the PID values generated with the control method described in~\cite{Ota2019}
in order to ensure precise position control 
for each end-effector to follow a feedback trajectory and remain within a specified range. 
The control input for robots was
joint angle vector which is the function of action-velocities emitted from RL
controller by observing feedback from simulation and real motor values in
handshaking protocol supplied by Robotis{\textregistered}. 
Special care was taken when mapping the joint angles in a structured manner,
complying with the kinematics and dynamics of the manipulator,
in order to avoid the self-collision. 
Finally, to quantitatively validate the effectiveness of our BPM methodology,
we designed three malfunction modes: 
1) unresponsive joints,
2) joints with constant offset angle,
and
3) Joints with random precision. 
To increase the difficulty level of BPM learning,
for each malfunction mode, 
we have changed the number of faulty joints,
i.e., malfunction degree, 
from 1 to 4,

\subsection{Simulation Environment}

Our simulation setting consists of a high-fidelity robot model of 
our physical 8-DOF robotic arm,
and a virtual environment closely resembling our physical setting. 
Specifically,
in order to replicate our physical arm for the simulation setting, 
we modelled our 8-DOF robotic arm with CAD software 
through incorporating various modular design files of actuators, brackets,
gripper, and base plate in our assembly, as depicted in Fig.~\ref{fig:approach}(d). 
Our simulation profile cloned all physical properties and inertial properties 
with high fidelity through exporting the CAD designs to the simulation profile. 
The inter process communication (IPC) was built upon
Robot Operating System (ROS) ecosystem to handle parallel executions and
register low latency feedback coming from simulation, real hardware, and the BPM
controller. 
Finally, each agent was spawned as an independent ROS node while its
communication was synced at the minimum rate of 20Hz
that ensures all independent ROS nodes to maintain their
handshaking message passing protocol. 
%


\subsection{Experimental Procedure and Parameters}

In our experiments, 
our robotic agent,
an 8-DOF hyper-redundant robotic arm,
started with a motion policy, $\pi_0$,
pre-trained assuming no faulty joint
with the state-of-the-art deep RL method with continuous control 
as in~\cite{Lillicrap2016,Ota2019}.
Specifically, 
we used $\epsilon$-greedy algorithm to compute 1-step lookahead value function,
defined our reward function as quaternion difference between expected pose 
and terminal pose of an agent. 
After $\pi_0$ was learnt, we injected different modes of malfunction with different degrees.

We now define key learning variables for our BPM.
The state $s$ in our experiments is defined as 8$\times$2$\times$2 matrix consisting of joint angles and pose variables containing quaternion orientations and end-effector coordinates.
The action $a$ is defined as a vector of 
8 parametric joint angles. 
A reward $r$ is computed as an absolute pose difference. 
Finally, each RL episode terminates if either of the following two condition is true: (a) the agent's pose is
sufficiently closer to the goal (desired) pose, 
and (b) the agent exhausts its allowed
number of steps for one episode.



\section{Experimental Setup and Results Analysis}
\label{sec:res}

\begin{figure*}[htbp]
\centering
   \includegraphics[width=\linewidth]{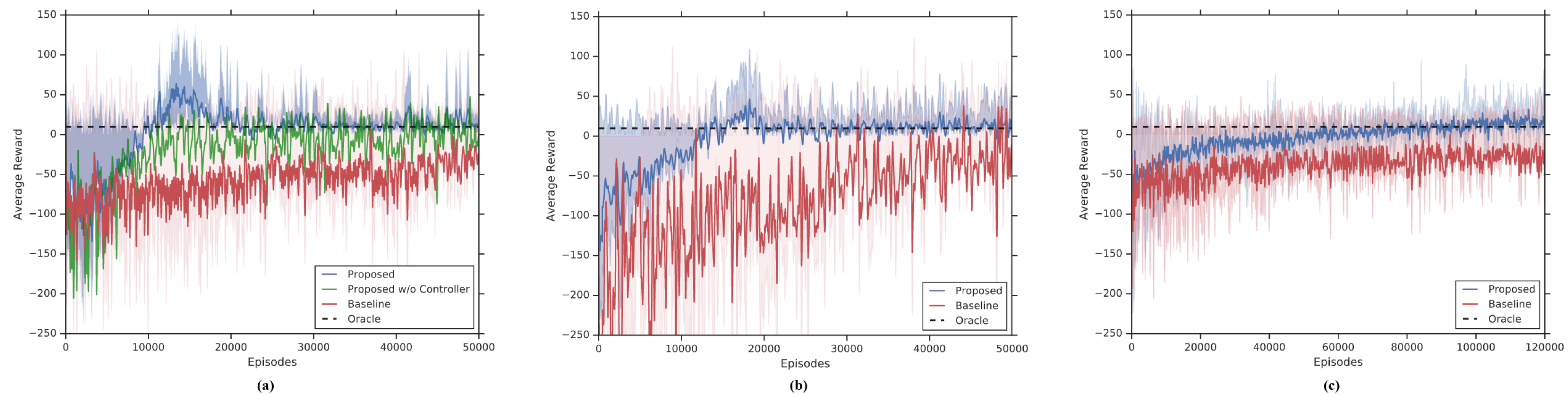}
\caption{
Comparison of learning convergence between BPM and the baseline DDPG for 
three malfunction modes: 
1) unresponsive joints,
2) joints with constant offset angle,
and
3) Joints with random precision. 
}
\label{fig:sim}
\end{figure*}

Given the experimental setup described in Section~\ref{sec:system}, 
we define the success of our robotic as 
the accuracy of reaching objective locations in the target area.
Our experiments seek to investigate the following: 

\begin{enumerate}

\item Does improve the learning efficiency when subjected to a joint malfunction?

\item How does the learning effectiveness of BPM vary with different degrees of malfunction, i.e., different number of faulty joints?

\item How does various key algorithmic components of BPM impact of its learning convergence?
Specifically, if removing the policy bias filtering, how adversely will the learning performance of 
our BPM methodology be impacted?

\end{enumerate}


In the rest of this section, 
after describing our experimental baseline, 
we present both experimental and simulation results to 
answer the above three questions quantitatively.


\subsection{Experimental Baseline}

To validate our BPM algorithm, 
we choose a recent study based on the Gaussian DQN method~\cite{robots-adapt-like-human-nature}
as our baseline, 
which has an almost identical settings and problem formulation.
The key idea of our baseline study can be
described as an incremental control policy learning termed as 
self-modelling in~\cite{robots-adapt-like-human-nature}, 
which is actually a Gaussian distribution of its MDP learnt offline. 
Specifically, the agent in~\cite{robots-adapt-like-human-nature} 
exploits its self-model when choosing
actions, and if a disagreement is detected between self-model and observation,
our baseline first infers the damage by choosing motor actions and measuring
their consequences as feedback to agent. Next, our baseline updates it's
self-model by making incremental changes to the existing Gaussian Process (GP)
until it reaches to optimal representation of observation.
This method has demonstrated significant performance in dense environment
spaces like quadrupedal agents \cite{quadrupedal-robot}, whose action space can
be discretized with fine granularity. However for agents in continuous spaces,
the baseline would encounter two problems:  
(a) it would be practically
infeasible to provide the self-model GP of a continuous space, 
(b) updating the
entire GP in each episodic setting to match with current observation would be
quite computationally expensive.
As discussed in Section~\ref{sec:approach}, 
our BPM methodology aims at continuous control 
by combining behavior learning baseline
with incremental actor-critic setting~\cite{incremental-actor-critic}. Instead
of providing a GP model, BPM provides the agent with 
the behavior ensemble of its model that has abstract information of state-action values. The agent does not
update the ensemble, instead it emits a bias hyperparameter from critic
network, by comparing observation and beliefs at each episode.


\subsection{BPM Learning under Different Malfunction Modes}

To thoroughly
validate our BPM methodology, we consider three different malfunction modes:
1) unresponsive actuators that mimic the complete failure of one or multiple
joints randomly selected during operation,
2) permanent angular offsets of 45 degrees for one or more joints,
and 3)
joints with randomly varying angles within a bound of $\pm$10 degrees.
As shown in 
Fig.~\ref{fig:sim},
our BPM algorithm, once reaching
to the oracle point, exhibits much lower variance around the oracle when compared with our baseline's
results. Note that the variance span for both methods are shown with lightly shaded
regions.
Additionally, since one of the key innovations of our BPM algorithm is its filtering mechanism,
the step of accept or reject in Algorithm~\ref{algo:pick}, that 
effectively controls the BPM's bias induction,  
we validate its importance by presenting 
the learning results without such bias induction with the green curve.
in Fig.~\ref{fig:sim}(a). 
As expected, the BPM without the final bias filtering demonstrates significantly lower learning rates
than the complete BPM,
albeit still  
better than the baseline.
This is largely because that the behavior ensemble under consideration 
does not carry complete representation of current observations and it is likely that,
in some states, it
can mislead the learning process to a local minima, resulting in sub-optimal
returns.

In Fig.~\ref{fig:sim}(b), 
we illustrate how well our BPM methodology will adapt to the ``stuck-at" error of servo joints. 
For our BPM setting, a joint
offset can be seen as an unknown but absolute translation or shift in a joint
angle value. 
With a constant 45-degree angular offset in one randomly chosen joint,
our BPM learns
this offset first with higher exploration (notice the high onset variance) until its first success
and drastically reduces its exploration space around the oracle. 
One way to intuitively understand the effectiveness of our BPM 
is to draw an analogy between our case with the classic 
multi-arm bandit learning using Thompson sampling~\cite{multi-arm-bandit}.
Fundamentally, our BPM algorithm treats its action space as contextual bandits with dynamic
distributions and uses Bayes rule to update those distributions incrementally.
Therefore, if any distribution encounters a permanent change,
our BPM, instead of
nudging the distribution, iteratively tunes the bias parameter until its
posterior matches with its observation.
In contrast,
our baseline method seemed to
lag behind throughout entire training period but still shows slowing improvement in its learning 
performance.


Finally in Fig.~\ref{fig:sim}(c), 
we study the impact of uncertainty in sensorimotor control upon the BPM learning.
Modeling the error profiles of robot actuators has been
extensively studied~\cite{bayesian-sensory}.
We instead focus on investigating 
if our BPM agent could adapt to random
sensorimotor noise without providing a detailed error profile.
As such, we artificially injected a uniformly sampled $\pm$10-degree angular noise  
in a randomly selected servo motor.
Fig.~\ref{fig:sim}(c) shows that, to combat such angular imprecision,
our BPM algorithm needed significantly more epochs for its pre-defined oracle level.
However, relative to the baseline, our BPM 
clearly demonstrated very controlled
variance (shaded region around average reward signal), which signifies 
the effectiveness of both controlled bias induction and average-based Monte Carlo sampling instead of
naive value estimates, which matches with our understanding that 
the naive value estimation, being greedy in nature,
works better for short-term rewards but will show high variance in long term.


\subsection{Impact of Malfunction Degree upon BPM Learning}

\begin{figure}[htbp]
    \begin{center}
        \includegraphics[width=\linewidth]{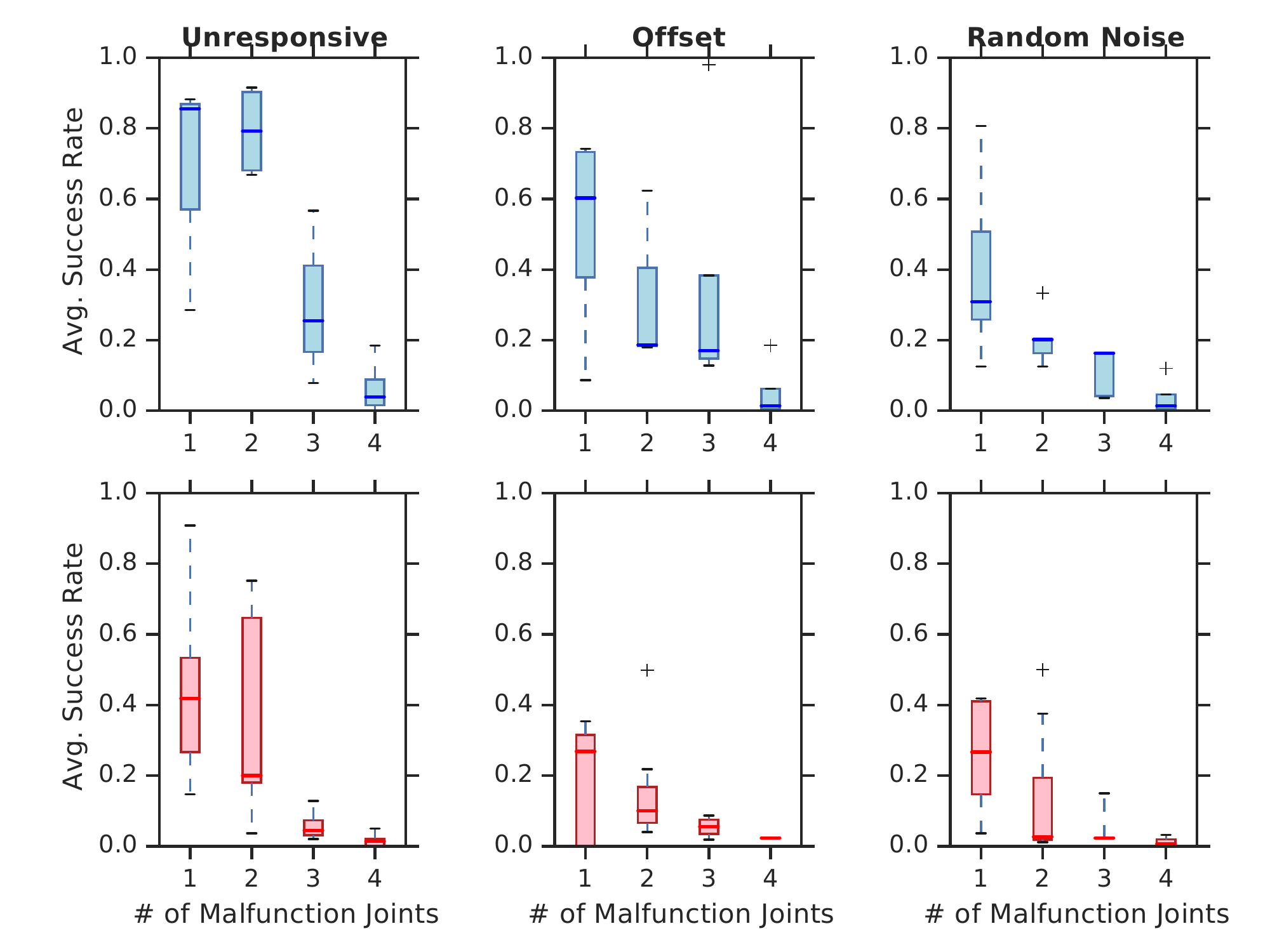}
        \caption{
Top row: Average success rate of proposed BPM approach, and  
Bottom row: Average success rate of baseline approach for three malfunction categories.
}
        \label{fig:boxplots}
    \end{center}
\end{figure}

We now investigate the effectiveness of our BPM methodology for different 
malfunction modes with various degrees. For each malfunction degree, We
randomly selected the locations of faulty joints.
In Fig.~\ref{fig:boxplots}, we have plotted our experimental results in six subplots,
in each of which, the shaded boxes show the confidence interval around median of success rates in
different malfunction configurations for all test cases,
while the whiskers show the errors computed by bootstrapping and extrapolating error bounds. 
Additionally, the thick line shows the median success ratio among an entire batch of randomly-seeded
samples, where all the $+$ points depicted the outliers that lie beyond the bootstrapping
bounds. 

Overall, for each malfunction category, our BPM method (blue boxes) has shown 
significant improvements over our baseline (red boxes) in terms of average success rates.
Specifically, for each category, as the malfunction degrees 
increase from 1 to 4, not surprisingly, the average success rate drops noticeably.
Interestingly, only for the category of unresponsive joints,
our BPM method actually achieved a slightly higher success rate when the number of broken joints 
changed from 1 to 2. One possible explanation is that two damaged joints actually reduced the DOF by 2, 
hence barely removing the hyper-redundancy.    
Our BPM methodology performed comparatively lower for the category of angular imprecision when compared with the other two categories, which may be due to the fact that the 3rd malfunction mode is significantly stochastic and highly non-linear.

\section{Conclusion}

The present research aims at 
adaptively ``morph" an existing robotic policy 
when sudden mechanical malfunctions occur, either statically or stochastically.
Our proposed BPM method utilizes Bayesian inference for morphing parametric
perturbations in order to bias gradient updates during policy optimization. 
The bias
would influence gradient directions based on maximum likelihood of prior
distribution correlating with on-going learning representation, maximizing
expected return. 
The major contributions of our BPM methodology include: 

\begin{itemize}

\item Instead of learning transition probabilities, our BPM algorithm 
exploits existing parametric distributions in order to estimate values incrementally, hence reducing the overall sampling complexity. 

\item Our BPM methodology  
provides a framework to guide action-selection (exploration/exploitation)
search by inducing a biasing kernel of uncertainty and expected returns, 
therefore achieving
more generalized policy representation than the standard reward-based policy gradient
methods such as DDPG and Bayesian Policy Reuse. 

\item 
Our proposed BPM method was demonstrated on  both 
simulated and real working environments using a high-fidelity simulator and
an research-grade 8-DOF robotic manipulator. 
We compared our adapted policy with BPM against a baseline controller
trained with the widely implemented DDPG algorithm augmented with continuous control scheme.
Our extensive experimentation has clearly demonstrated a significant boost in 
survivability for our robotic arm.

\end{itemize}

In future research, we would like to investigate the 
proposed BPM algorithm for robotic agents equipped with multi-modal sensors, thus 
perceiving their states and actions autonomously through sensory data fusion.





\bibliographystyle{IEEEtran}
\bibliography{bib.bib}

\end{document}